%% file: neurips_2026.tex
\documentclass{article}

\usepackage[preprint]{neurips_2026}

\usepackage[utf8]{inputenc} 
\usepackage[T1]{fontenc}    
\usepackage{hyperref}       
\usepackage{url}            
\usepackage{booktabs}       
\usepackage{amsfonts}       
\usepackage{nicefrac}       
\usepackage{microtype}      
\usepackage{xcolor}         
\usepackage{multirow}
\usepackage{graphicx}
\usepackage[table]{xcolor}
\usepackage{adjustbox}
\usepackage{subcaption}
\usepackage{algorithm}
\usepackage{algpseudocode}
\usepackage{tikz}
\usepackage{xcolor}
\usepackage{fontawesome5}
\usepackage{enumitem}
\usepackage{amssymb}
\usepackage{pgfplots}
\usepackage{wrapfig}
\usepackage{amsmath}
\usepackage{tcolorbox}
\tcbuselibrary{breakable}
\usepackage{listings}
\usepackage{tcolorbox}
\tcbuselibrary{breakable}
\usepackage{fvextra}
\usepackage{fancyvrb}
\usetikzlibrary{positioning, arrows.meta, calc, fit,
                shapes.geometric, shapes.misc, backgrounds,
                decorations.pathreplacing}

\definecolor{SearchHighlight}{RGB}{255,243,205}

\title{LLMs Improving LLMs: \\ Agentic Discovery for Test-Time Scaling}

\author{%
\small
Tong Zheng\textsuperscript{1},
Haolin Liu\textsuperscript{2},
Chengsong Huang\textsuperscript{3},
Huiwen Bao\textsuperscript{},
Sheng Zhang\textsuperscript{1},
Rui Liu\textsuperscript{1},
Runpeng Dai\textsuperscript{4},
\\[-0em]
\bf
Ruibo Chen\textsuperscript{1},
Chenxi Liu\textsuperscript{1},
Tianyi Xiong\textsuperscript{1},
Xidong Wu\textsuperscript{5},
Hongming Zhang\textsuperscript{6}
Heng Huang\textsuperscript{1}
\\[-0em]
\small
\textsuperscript{1}UMD,
\textsuperscript{2}UVA,
\textsuperscript{3}WUSTL,
\textsuperscript{4}UNC,
\textsuperscript{5}Google,
\textsuperscript{6}Meta
}

\begin{document}

\maketitle

\begin{abstract}
Test-time scaling (TTS) has become an effective approach for improving large language model performance by allocating additional computation during inference. 
However, existing TTS strategies are largely hand-crafted: researchers manually design reasoning patterns and tune heuristics by intuition, leaving much of the computation-allocation space unexplored. 
We propose an environment-driven framework, \textbf{AutoTTS}, that changes what researchers design: from individual TTS heuristics to environments where TTS strategies can be discovered  automatically. 
The key to AutoTTS lies in environment construction: the discovery environment must make the control space tractable and provide cheap, frequent feedback for TTS search.
As a concrete instantiation, we formulate width--depth TTS as controller synthesis over pre-collected reasoning trajectories and probe signals, where controllers decide when to branch, continue, probe, prune, or stop and can be evaluated cheaply without repeated LLM calls.
We further introduce beta parameterization to make the search tractable and fine-grained execution trace feedback to improve discovery efficiency by helping the agent diagnose why a TTS program fails.
Experiments on mathematical reasoning benchmarks show that the discovered strategies improve the overall accuracy--cost tradeoff over strong manually designed baselines. 
The discovered strategies generalize to held-out benchmarks and model scales, while the entire discovery costs only \textbf{\$39.9} and \textbf{160 minutes}. Our data, and code will be open-source at \url{https://github.com/zhengkid/AutoTTS}.
\end{abstract}

\input{Main/intro_auto_tts_v2}

\input{Main/problem_setup}

\input{Main/method}

\input{Main/experiment_setup}

\input{Main/experiment}

\input{Main/related_work}

\input{Main/Conclusion}


{
\small
\bibliographystyle{unsrt}
\bibliography{neurips_2026}
}


\input{Main/appendix}


\end{document}

%% file: Main/intro_auto_tts_v2.tex
\begin{figure*}[h]
  \centering
  \includegraphics[width=\linewidth]{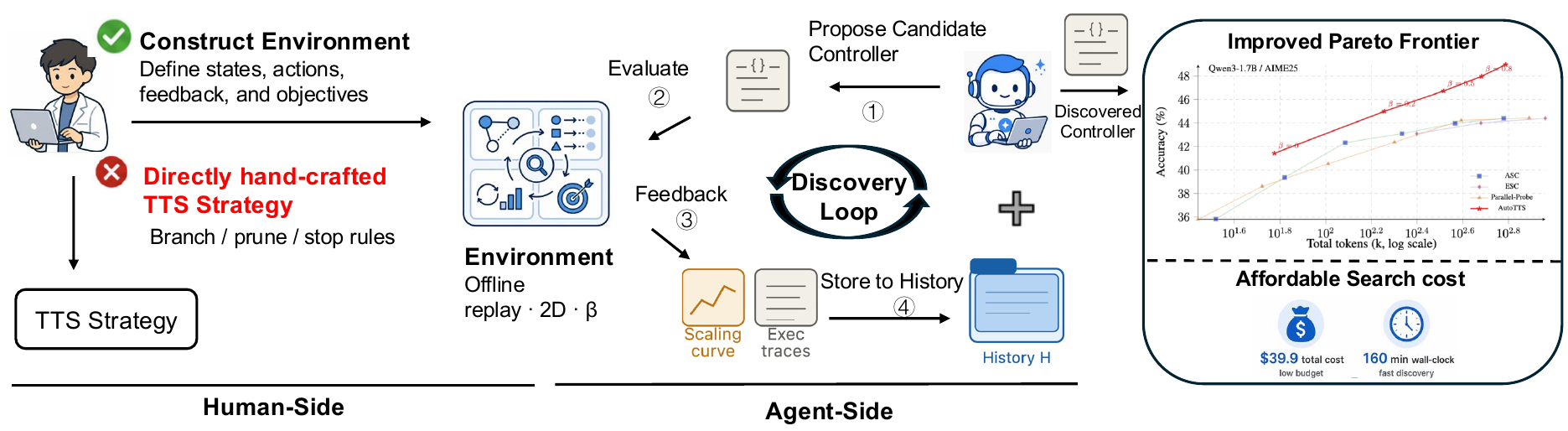}
   \caption{
\textbf{Overview of our Auto-TTS framework.}
Unlike the traditional workflow of manually designing TTS strategies, Auto-TTS shifts the human role from directly hand-crafting branching, pruning, and stopping heuristics to constructing environments by defining states, actions, feedback, and objectives. Given the constructed environment, an explorer LLM iteratively proposes candidate controllers, evaluates them in the offline replay environment, receives feedback from scaling curves and execution traces, and uses the accumulated history to refine future proposals. The right panel shows an example evaluation on Qwen-1.7B and AIME25, where the discovered controller improves the accuracy--cost Pareto frontier over hand-crafted baselines with an affordable one-time search cost.
}
  \label{fig:overview}
\end{figure*}
\section{Introduction}

\begin{wrapfigure}[25]{r}{0.5\textwidth}
  \vspace{-2em}
  \centering
  \includegraphics[width=0.5\textwidth]{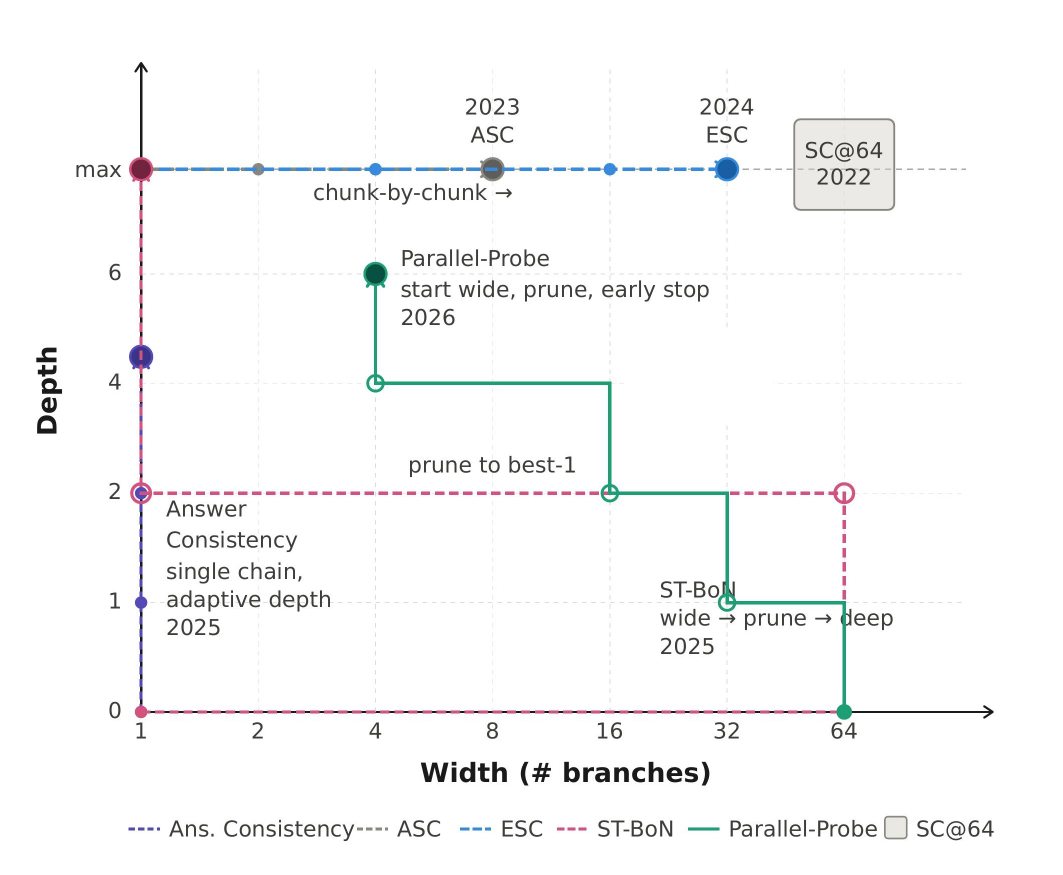}
  \caption{Existing TTS algorithms as special cases of the width--depth control space. Each algorithm traces a distinct path: \textsc{SC@64}~\cite{wang2022self} occupies a fixed full-budget corner; \textsc{ASC}~\cite{aggarwal2023let} and \textsc{ESC}~\cite{li2024escape} adapt only along the width axis at max depth; \textsc{Answer Consistency}~\cite{liu2025answer} adapts only along the depth axis on a single chain; \textsc{ST-BoN}~\cite{wang2025sampling} expands wide, prunes to one branch, then deepens; \textsc{Parallel-Probe}~\cite{zheng2026parallel} starts wide and progressively prunes while deepening.}
  \label{fig:tts_2d_space}
  \vspace{-1em}
\end{wrapfigure}

Test-time scaling (TTS)~\cite{snell2024scaling,brown2024large, muennighoff2025s1} has emerged as a powerful paradigm for improving large language model performance by allocating additional computation during inference. However, performance depends not just on the amount of computation used, but on how it is allocated~\cite{snell2024scaling} and existing strategies for doing so are largely hand-crafted: researchers manually hypothesize heuristics for when to branch, deepen, probe, prune, or stop reasoning trajectories~\cite{zhao2025majority,wen2025parathinker,wang2026not,zheng2026parallel,tu2025deepprune,zhang2025alphaone}, implement them, and tune thresholds by intuition.

Looking back at the development of TTS strategies reveals a valuable perspective. Although existing methods differ substantially in form, many of them can be interpreted as manually specified policies within some underlying computation-allocation space. A simple example is the width--depth space, where width denotes how many reasoning branches are explored and depth denotes how far each branch is developed, as illustrated in Figure~\ref{fig:tts_2d_space}. Under this view, several representative methods correspond to different trajectories through the space: some expand width by sampling more reasoning branches~\cite{wang2022self,aggarwal2023let,li2024escape}; some increase depth by extending reasoning trajectories~\cite{muennighoff2025s1,zhang2025alphaone}; and others introduce adaptive stopping, pruning, or selection rules to move through the space more selectively~\cite{fu2025deep,tu2025deepprune,liu2025answer,zheng2026parallel}. Notably, this perspective is not intended to reduce all TTS algorithms to a two-dimensional abstraction, as many methods involve richer structures such as tree search~\cite{yao2023tree,inoue2025wider} or verifier-guided refinement~\cite{snell2024scaling,wang2024math,luo2024improve}. 
Rather, the case of width--depth space reveals that many TTS strategies can be seen as hand-designed special cases within a structured control space.

\textbf{This perspective suggests a fundamental reframing of the problem.}  In this work, we propose \textsc{AutoTTS}, an environment-driven paradigm for automatic TTS strategy discovery (Figure \ref{fig:overview}). Instead of hand-crafting individual branching, pruning, and stopping heuristics, AutoTTS shifts the human role to constructing discovery environments, where humans define the control space through states, actions, feedback, and objectives, and agents search within this space for effective allocation policies.

As a proof-of-concept instantiation, we formulate width--depth test-time scaling (Figure \ref{fig:tts_2d_space}) as controller synthesis in an offline replay environment. 
For each problem, we pre-collect reasoning trajectories and intermediate probe signals, allowing a controller to replay decisions over when to branch, continue, probe, prune, or stop. 
The controller observes active branches, their depths, revealed probe outputs, and the remaining budget, and is evaluated by the resulting accuracy--cost tradeoff. Since evaluation reuses pre-collected trajectories, candidate controllers can be assessed cheaply and deterministically without repeatedly invoking the base LLM. However, effective discovery faces two additional challenges: automatically discovered controllers tend to introduce excessive hyperparameters that makes the search space large and difficult to navigate within a limited number of rounds, and scalar accuracy--cost feedback alone is not sufficient to tell the explorer why a controller fails. 

To make controller search tractable,
we introduce beta parameterization, where each controller exposes only one scalar trade-off parameter $\beta$ and derives all internal hyperparameters deterministically from it, reducing overfitting to the search set. 
Additionally, we address the feedback issue by logging execution traces that expose how each controller allocates computation over time, enabling the explorer to diagnose failure modes and propose targeted improvements.

Experiments on mathematical reasoning benchmarks show that AutoTTS discovers controllers that improve the accuracy–cost Pareto frontier over strong hand-crafted baselines. The discovered controllers generalize from the search benchmark to held-out benchmarks and across model scales, while the entire discovery process remains affordable due to fixed replay. \textbf{These results suggest that environment-driven discovery offers a scalable and reusable alternative to manually designing TTS strategies, and that the right place to invest human effort is in environment design, not strategy design.}

%% file: Main/problem_setup.tex
\section{Test-Time Scaling as Algorithmic Search}
\label{sec:problem_setup}


We consider adaptive test-time algorithms that allocate inference budget across
multiple reasoning branches. For each question \(q\), a controller may create,
extend, probe, and prune branches before aggregating the explored prefixes into a
final answer, covering strategies such as best-of-\(N\), self-consistency, early
stopping, and adaptive branching. Each branch \(i\) produces prefixes
\(z_{i,1},z_{i,2},\ldots\), where \(z_{i,k}\) is obtained after \(k\) fixed-length
generation intervals. Each prefix induces an intermediate answer \(\omega_{i,k}\) (i.e., the
answer that would be produced from the current prefix),
which is observed only if the controller explicitly takes a probing action at that branch.

Since we consider a unit generation as an interval with a fixed token length, we
measure computation in units of intervals. At decision step \(t\), let
\(m_t\in\mathbb Z_{\ge 0}\) denote the number of branches instantiated so far,
and use the convention \([m_t]=\{1,\ldots,m_t\}\), with \([0]=\emptyset\). The
state is \(s_t=(q,m_t,I_t,\ell_t,Z_t,\Omega_t)\), where \(q\in\mathcal Q\) is
the question, \(I_t\subseteq [m_t]\) is the set of currently active branches,
\(\ell_t=(\ell_{t,i})_{i\in[m_t]}\) records the current depth of every
instantiated branch, \(Z_t=(Z_{t,i})_{i\in[m_t]}\) records the generated
prefixes, and \(\Omega_t\) is the set of probe feedback revealed so far. For
every instantiated branch \(i\in[m_t]\), \(\ell_{t,i}\ge 1\) denotes how many
fixed-length generation intervals have been generated on that branch, and
\(Z_{t,i}=(z_{i,1},\ldots,z_{i,\ell_{t,i}})\) contains the prefixes generated on
branch \(i\) up to its current depth. For a pruned branch
\(i\in[m_t]\setminus I_t\), \(\ell_{t,i}\) and \(Z_{t,i}\) record the depth and
generated prefixes at which it was pruned. The revealed feedback set satisfies
\(\Omega_t\subseteq\{(i,k,\omega_{i,k}): i\in[m_t],\ 1\le k\le \ell_{t,i}\}\).
Unrevealed probe outputs are not part of the controller state. The computation
cost of a state is \(\mathrm{Cost}(s_t)=\sum_{i=1}^{m_t}\ell_{t,i}
+\kappa_{\mathrm{probe}}|\Omega_t|\), where \(\kappa_{\mathrm{probe}}\ge 0\) is
the cost of reading one probe signal. In settings where probing is treated as
free relative to generation, we set \(\kappa_{\mathrm{probe}}=0\).

Given \(s_t=(q,m_t,I_t,\ell_t,Z_t,\Omega_t)\), the admissible action set is
\(\mathcal A(s_t)=\{\texttt{BRANCH}\}\cup\{\texttt{CONTINUE}(i):i\in I_t\}\cup
\{\texttt{PROBE}(i):i\in I_t,\ \nexists \omega\ \mathrm{s.t.}\
(i,\ell_{t,i},\omega)\in\Omega_t\}\cup\{\texttt{PRUNE}(i):i\in I_t\}\cup
\{\texttt{ANSWER}\}\). Here \(\texttt{BRANCH}\) creates a new branch \(m_t+1\)
and advances it from the question to the end of the first interval.
\(\texttt{CONTINUE}(i)\) advances branch \(i\) by one fixed-length generation
interval. \(\texttt{PROBE}(i)\) reveals the current probe signal
\(\omega_{i,\ell_{t,i}}\) without advancing the branch. \(\texttt{PRUNE}(i)\)
removes branch \(i\) from the active set, while keeping its current information.
\(\texttt{ANSWER}\) terminates inference and invokes the aggregation rule. An
aggregation rule \(\operatorname{Agg}\) takes a state as input and outputs the
final answer.

Formally, the initial state is
\(s_0=(q,0,\emptyset,\emptyset,\emptyset,\emptyset)\). If
\(a_t=\texttt{BRANCH}\), a new branch \(m_t+1\) is instantiated and advanced to
the first probe point, producing prefix \(z_{m_t+1,1}\). Then
\(m_{t+1}=m_t+1\), \(I_{t+1}=I_t\cup\{m_t+1\}\),
\(\ell_{t+1}=(\ell_t,1)\), \(Z_{t+1}=(Z_t,(z_{m_t+1,1}))\), and
\(\Omega_{t+1}=\Omega_t\). If \(a_t=\texttt{CONTINUE}(i)\), branch \(i\) is
advanced by one interval, producing prefix \(z_{i,\ell_{t,i}+1}\). Then
\(m_{t+1}=m_t\), \(I_{t+1}=I_t\), \(\Omega_{t+1}=\Omega_t\),
\(\ell_{t+1,j}=\ell_{t,j}+\mathbf 1\{j=i\}\) for all \(j\in[m_t]\),
\(Z_{t+1,i}=(Z_{t,i},z_{i,\ell_{t,i}+1})\), and \(Z_{t+1,j}=Z_{t,j}\) for all
\(j\neq i\). If \(a_t=\texttt{PROBE}(i)\), then \(m_{t+1}=m_t\),
\(I_{t+1}=I_t\), \(\ell_{t+1}=\ell_t\), \(Z_{t+1}=Z_t\), and
\(\Omega_{t+1}=\Omega_t\cup\{(i,\ell_{t,i},\omega_{i,\ell_{t,i}})\}\). If
\(a_t=\texttt{PRUNE}(i)\), then \(m_{t+1}=m_t\),
\(I_{t+1}=I_t\setminus\{i\}\), \(\ell_{t+1}=\ell_t\),
\(Z_{t+1}=Z_t\), and \(\Omega_{t+1}=\Omega_t\). If
\(a_T=\texttt{ANSWER}\) at time \(T\), the episode terminates and the final answer
is produced by \(\hat y=\operatorname{Agg}(s_T)\).

Our goal is to find a code-defined policy \(\pi\) that maps a state and a
hyperparameter \(\beta\) to a distribution over admissible atomic actions  $\pi(\cdot\mid s,\beta)\in\Delta(\mathcal A(s))$.
Here, \(\beta\) specifies the tunable hyperparameters of the algorithm. We also
allow each controller to include its own terminal aggregation rule
\(\operatorname{Agg}_{\pi,\beta}\). For a question \(q\), let
\(\tau=(s_0,a_0,s_1,a_1,\ldots,s_T)
    \sim P_{\pi,\beta}(\cdot\mid q)
\)
denote the execution trajectory induced by running \((\pi,\beta)\) from the
initial state \(s_0=(q,0,\emptyset,\emptyset,\emptyset)\) until
\(\texttt{ANSWER}\) is selected at time \(T\). The trajectory distribution
\(P_{\pi,\beta}(\cdot\mid q)\) includes the randomness from the policy, the
generation process, and any stochastic probe signals. At the terminal state
\(s_T\), the final answer and computation cost are
\(\hat y_{\pi,\beta}(\tau)
    =
    \operatorname{Agg}_{\pi,\beta}(s_T),
    C(\tau)=\operatorname{Cost}(s_T).
\)
For a task distribution \(\mathcal D\) over question-answer pairs \((q,y)\), we
choose \((\pi,\beta)\) to maximize accuracy while controlling computation cost.
With trade-off parameter \(\gamma\), the objective is
\begin{align*}
    \max_{(\pi,\beta)}
    \mathbb E_{(q,y)\sim\mathcal D,\ \tau\sim P_{\pi,\beta}(\cdot\mid q)}
    \left[
        \mathbf 1\{\hat y_{\pi,\beta}(\tau)=y\}
        - \gamma C(\tau)
    \right].
\end{align*}

The discovery loop searches over code-defined controllers. Each candidate is run
on every question to obtain an execution trajectory \(\tau\); we compare its final
answer with the ground truth and record its computation cost. The resulting
execution histories are stored in memory and used to guide subsequent rounds of
policy search. Finally, we output the code-defined policy with the best
accuracy--cost trade-off.

%% file: Main/method.tex
\section{AutoTTS: Environment-Driven Discovery}
\label{sec:method}

We instantiate \textsc{AutoTTS} as a concrete discovery pipeline for the objective in Section~\ref{sec:problem_setup}.
The key challenge is making the search over code-defined policies $\pi(\cdot \mid s, \beta)$ tractable:
evaluating a candidate policy online requires generating token interval $z_{i,k}$ and probing answer $\omega_{i,k}$ on demand,
which is prohibitively expensive at search time.
We address this challenge through three complementary design choices:
an \textbf{offline replay environment} that eliminates repeated LLM calls during evaluation (Section~\ref{sec:method:env}),
\textbf{beta parameterization} that prevents overfitting to the search set (Section~\ref{sec:method:beta}),
and \textbf{execution trace feedback} that enables the agent to diagnose failure modes
rather than relying on scalar outcomes alone (Section~\ref{sec:method:loop}).

\subsection{Replay Environment Construction}
\label{sec:method:env}
The central challenge in evaluating candidate controllers online is that each evaluation requires invoking the base LLM to generate reasoning trajectories on demand, which is prohibitively expensive at search time. To address this, we construct an offline replay environment that moves all LLM calls prior to the discovery process, making controller evaluation deterministic and cheap.

\paragraph{Offline data collection.}Following the data collection protocol of Parallel-Probe~\cite{zheng2026parallel}, for each question $q\in\mathcal{Q}$, we pre-collect $N$ independent reasoning trajectories from the base LLM, each segmented into fixed-length intervals of $\Delta$ tokens. This directly instantiates the branch prefixes $z_{i,1}, z_{i,2}, \ldots$ and probe signals $\omega_{i,1}, \omega_{i,2}, \ldots$ introduced in Section~\ref{sec:problem_setup}, with all data stored offline before any discovery begins. Each controller decision is then executed against this pre-collected data rather than invoking the LLM, making repeated controller evaluation affordable.

\paragraph{Evaluation via offline replay.}  To evaluate each controller for a given $\beta$, we run it on 
each question in $\mathcal{Q}_{\mathrm{search}}$: at each state $s_t$ it 
selects an action from $\mathcal{A}(s_t)$ and advances until 
\textsc{Answer} action is taken. Because all branch prefixes and probe signals 
are pre-collected offline, each action reads deterministically from 
this stored data rather than invoking the LLM---for instance, a 
\textsc{Probe} action on branch $i$ at depth $k$ simply retrieves the 
pre-collected signal $\omega_{i,k}$ at zero generation cost. This makes 
the entire $\beta$ sweep affordable without any additional LLM calls.

\subsection{Discovery Loop}
\label{sec:method:loop}

We partition $\mathcal{Q}$ into a search set $\mathcal{Q}_{\mathrm{search}}$ and a held-out evaluation set $\mathcal{Q}_{\mathrm{eval}}$. \textsc{AutoTTS} discovers an effective controller through a multi-round loop: each round, an explorer LLM (Claude Code) reads the accumulated history $\mathcal{H}$ and proposes an improved controller by directly editing the code; the controller is then evaluated on $\mathcal{Q}_{\mathrm{search}}$ and the results are appended to $\mathcal{H}$.

\paragraph{Agent-driven proposal.} The explorer reads $\mathcal{H}$---which stores all previously proposed controller implementations, their accuracy--cost outcomes, and execution traces---and is prompted to analyse what went wrong in prior proposals, and propose a new controller that improves accuracy while reducing token usage (full prompt in Appendix~\ref{app:prompt}).

\paragraph{History Design.} While scalar outcomes such as accuracy and token usage provide a coarse signal of whether a proposed controller is good enough, they reveal little about \textit{why} a controller fails. To address this limitation, we augment the history with the full decision-making trajectories that the controller executed during the replay environment. For each round, we sweep across multiple betas and record the resulting scaling curve as the scalar component; the trajectory component then supplies fine-grained behavioral evidence, enabling the agent to diagnose failure modes and propose a more targeted controller in the next round. This design is consistent with the finding in~\cite{lee2026meta} that fine-grained execution feedback improves agentic discovery for harness engineering.

\paragraph{Controller selection.} After $R$ rounds, we select the 
controller and $\beta$ value that achieves the highest accuracy on 
$\mathcal{Q}_{\mathrm{search}}$.

\subsection{Beta Parameterization for Tractable Search}
\label{sec:method:beta}

In our preliminary experiments, we empirically find that agents tend to propose 
TTS controllers with a large number of hyperparameters, up to 10. With only five discovery rounds, navigating this high-dimensional space causes the agent to collapse onto extreme solutions—such as overly aggressive pruning thresholds—that happen to minimize token cost on the search set but fail to represent robust allocation strategies.

To mitigate this, we propose \textbf{beta parameterization}: each controller must 
expose only a single hyperparameter $\beta$ and implement a map function from 
$\beta$ to all internal hyperparameters. We further require this map to be monotone, 
such that larger $\beta$ corresponds to larger token budget. This collapses the 
search space to a one-dimensional sweep and prevents the agent from discovering 
sharp, search-set-specific thresholds. Notably, the map function is produced directly by the coding agent.

%% file: Main/experiment_setup.tex
\section{Experimental Setup}
\label{sec:experimental_setup}
\paragraph{Experimental Protocol.}
All experiments use offline replay environments, each built from a 
specific (model, benchmark) pair across four Qwen3 models (0.6B, 
1.7B, 4B, 8B)~\cite{yang2025qwen3}. Following Parallel-Probe~\cite{zheng2026parallel}, we pre-sample 128 reasoning trajectories per (model, problem) pair at temperature 0.7 with a probing interval of 500 tokens to construct the replay matrix. To reduce variance, each controller is evaluated 64 times independently by randomly sampling a subset of trajectories from the pre-sampled pool of 128, and the results are averaged. For discovery, we use 
AIME24 as $\mathcal{Q}_{\text{search}}$ and construct 
$\mathcal{E}_{\text{search}}$ as the union of AIME24 environments 
across all four models. The discovery loop runs for five rounds with 
Claude Code as the agent, and the final controller is selected as the one achieving the highest accuracy on $\mathcal{E}_{\text{search}}$. 
The discovered controller (Appendix \ref{app:controller}) is fixed and evaluated on held-out 
environments from AIME25 and HMMT25, which are never used during 
discovery or selection.

\paragraph{Baselines.}
To demonstrate the effectiveness of our proposed discovery framework, we compare the discovered algorithm with several representative handcrafted test time scaling methods. 1) \textbf{Self-Consistency (SC@64)~\cite{wang2022self}:} A vanilla parallel reasoning approach that first samples 64 reasoning trajectories and performs majority voting to obtain the final answer; 2) \textbf{ASC~\cite{aggarwal2023let}:} A parallel sampling approach that samples  trajectories one by one and stop until reaching a pre-defined threshold. We follow the original
setting with threshold 0.95; 3) \textbf{ESC~\cite{li2024escape}:} A chunk-based hybrid approach that generates trajectories in parallel and terminates early when answer stability is detected within a sliding window. We use a chunk size of 8 and 4) \textbf{Parallel-Probe~\cite{zheng2026parallel}:} A recent efficient parallel reasoning approach that leverages cross-branch information to dynamically decide when to stop reasoning, prune unpromising branches, or continue computation.

\paragraph{Metrics.}
We report both task accuracy and tokens that measure the total number of tokens consumed across all used branches.

%% file: Main/experiment.tex
\section{Results and Analysis}
\label{sec:results_and_analysis}
\subsection{Main Results}
\label{sec:main_results}
\input{Raw_Results/main}
Table~\ref{tab:main_results} shows the performance of both handcrafted baselines and the discovered controllers on AIME24 (search), AIME25 (held-out), and HMMT25 (held-out) across four models. We can see that the discovered controllers achieve a better accuracy--cost tradeoff in most settings. Generally, the discovered controller, optimized solely on AIME24, generalizes to held-out benchmarks AIME25 and HMMT25, outperforming all handcrafted baselines in three out of four models on average, and remaining competitive on Qwen3-8B (62.7 vs.\ 62.8 for SC@64). Furthermore, the controller discovered by AutoTTS achieves better accuracy--token tradeoffs. For example, when $\beta=0.5$, it reduces token consumption by approximately 69.5\% compared to SC@64 while maintaining on-par accuracy on average across all four models (45.3 vs.\ 45.2); when $\beta=1.0$, it pushes peak accuracy beyond all handcrafted baselines in 5 out of 8 cases.

\subsection{Accuracy--Cost Scaling Curves}
\label{sec:budget_scaling}

\input{Fig/scaling_curves}
Figure~\ref{fig:scaling_main} compares the accuracy--token scaling curves of discovered and handcrafted controllers on held-out benchmarks. For handcrafted baselines, we control the reasoning budget by varying the (maximum) number of sampled trajectories. For the discovered controller, we vary the parameter $\beta$ to control the budget. 
Overall, the discovered controller consistently achieves a stronger accuracy--efficiency frontier, as indicated by its corresponding curves being positioned above the other curves across all four settings. Additionally, another key observation is that our discovered controller does not simply reduce inference cost at a fixed accuracy level.

\input{Raw_Results/generalization}
By contrast, it can also push the attainable peak performance on the four settings, respectively. This indicates that the discovered controller can better identify noisy or unproductive branches and redirect computation toward branches that contribute useful reasoning signals.

\subsection{Generalization Beyond Qwen Models and Math Tasks}
\label{sec:generalization}
Table~\ref{tab:generalization_beyond_main} evaluates whether the discovered controller transfers beyond the main experimental setting.
We consider two targeted generalization scenarios: DeepSeek-R1-Distill-Llama-8B, a Llama-based model distilled from DeepSeek-R1~\cite{guo2025deepseek} on HMMT25, and a non-math benchmark (GPQA-Diamond~\cite{rein2023gpqa}) with Qwen3-1.7B. On DeepSeek-R1-Distill-Llama-8B with HMMT25, our discovered controller with $\beta=1$ achieves the best accuracy among all methods while also reducing total tokens from 985.7K under SC@64 to 533.9K. The lower-budget variant with $\beta=0.5$ further reduces the token cost to 279.0K, at the cost of a small accuracy drop. On GPQA-Diamond, the discovered controllers also remain competitive. Both $\beta=1$ and $\beta=0.5$ achieve comparable or slightly better accuracy than SC@64, while using substantially fewer tokens. In particular, the $\beta=0.5$ variant reduces the token cost from 510.0K to 151.0K, and also improves over ASC in both accuracy and token usage.
These results suggest that the discovered strategies are not tightly overfitted to the main Qwen-based math setting, and can transfer to both a different model family and a non-math reasoning benchmark.

\subsection{Ablation Study}
\label{sec:ablation}
We further conduct an ablation study to examine the key design choices including beta-controlled search space and history design, in our discovery framework. Table~\ref{tab:ablation} reports the results on AIME24, AIME25, and HMMT25. For each dataset, we report the average score across all four models.

\input{Raw_Results/ablation_studies}

Removing \textit{Beta Parameterization} leads to controllers with excessive free hyperparameters, creating a large search space that makes it easy to overfit to $\mathcal{E}_{\text{search}}$. This overfitting manifests as overly aggressive pruning and stopping thresholds that happen to work well on the search set but fail to generalize, resulting in a drastic token reduction (575.5K to 93.3K) accompanied by an accuracy drop from 53.1 to 49.0 on held-out benchmarks. By contrast, beta parameterization collapses all hyperparameters into functions of a single scalar $\beta$, and the discovery prompt further constrains this mapping to be monotone in $\beta$, 
which drastically reduces the effective search space and mitigates overfitting to the search set. In addition, we find detailed \textit{Execution Traces} are essential for controller discovery. Without this, the discovered controller achieves worse performance while allocating much more tokens.  This indicates that final acc/token numbers alone are insufficient to guide effective search. Execution traces expose intermediate decisions, such as which branches are expanded, pruned, or stopped, allowing the search agent to diagnose failure modes of previous controllers and reuse successful computation-allocation patterns. This is consistent with the finding in Meta-Harness~\cite{lee2026meta}.

\subsection{Efficiency Analysis}
\label{sec:efficiency}
\input{Raw_Results/cost_efficiency}
Figure~\ref{fig:overview} reports the overhead of the discovery stage. We summarize two key metrics: discovery cost and wall-clock time. Overall, the five-round discovery process costs only \$39.9 and takes 160 minutes, suggesting that our replay-based discovery framework is practical to run. Notably, the feasible search wall clock is largely attributed to our formulation of TTS search on fixed replay environment, which avoids repeated evaluation via LLM calls.

\subsection{Evolution of Discovery Process}
\label{sec:discovered_algorithm_analysis}

\input{Fig/self_evolving}

Figure \ref{fig:self_evolving} visualizes the discovery trajectory over multiple search rounds. Each point corresponds to the controller proposed at one discovery round, evaluated with $\beta=1$. The left panel reports the accuracy--cost trajectory on the AIME24 search set, while the right panel reports the corresponding performance on the held-out AIME25 and HMMT25 benchmarks.

The trajectory shows that the discovery process progressively corrects the accuracy--cost allocation. The initial controller is overly aggressive in reducing token usage, leading to relatively low accuracy. After observing this failure mode, the explorer increases the computation budget in later rounds, which substantially recovers accuracy. Subsequent rounds then make more fine-grained adjustments: some steps push efficiency by reducing cost with little accuracy loss, while others push accuracy by allocating more computation. Overall, the trajectory moves toward a better accuracy--cost frontier rather than monotonically optimizing a single metric. The final discovered controller also achieves strong performance on held-out benchmarks, suggesting the improvements generalize beyond the search set.

\subsection{Analysis of Discovered Controller}
We also analyze the discovered controller (Appendix~\ref{app:controller}). 
The discovered controller reveals four non-obvious mechanisms: trend-based 
stopping via EMA momentum, coupled width--depth control through a shared 
evidence signal, alignment-aware depth allocation, and conservative branch 
abandonment. Together, their joint design represents a level of coordinated 
complexity that would be difficult to arrive at through manual intuition 
alone. Full details are provided in Appendix~\ref{app:controller}.

%% file: Raw_Results/main.tex
\begin{table*}[t]
\centering\small
\definecolor{trainColor}{rgb}{1.0,0.93,0.78}
\caption{Accuracy and total tokens. AIME24 (\colorbox{trainColor}{\strut search}) is used for controller discovery; AIME25 and HMMT25 are held-out.}
\vskip 0.1in
\label{tab:main_results}
\begin{adjustbox}{width=\textwidth}
\begin{tabular}{l l >{\columncolor{trainColor}}c >{\columncolor{trainColor}}c c c c c c c}
\toprule
Method & Type & \multicolumn{2}{>{\columncolor{trainColor}}c}{AIME24\,(search)} & \multicolumn{2}{c}{AIME25\,(held-out)} & \multicolumn{2}{c}{HMMT25\,(held-out)} & \multicolumn{2}{c}{Avg. (held out)} \\
\cmidrule(lr){3-4}\cmidrule(lr){5-6}\cmidrule(lr){7-8}\cmidrule(lr){9-10}
 &  & Acc. $\uparrow$ & Tokens $\downarrow$ & Acc. $\uparrow$ & Tokens $\downarrow$ & Acc. $\uparrow$ & Tokens $\downarrow$ & Acc. $\uparrow$ & Tokens $\downarrow$ \\
\midrule
\rowcolor{white}\multicolumn{10}{l}{\textit{Base Model: Qwen3-0.6B}} \\
\midrule
SC @ 64 & Handcrafted & 21.4 & 1008.6k & 28.9 & 890.5k & 18.1 & 937.8k & 23.2 & 914.2k \\
ASC & Handcrafted & 21.4 & 805.5k & 28.9 & 653.8k & 18.1 & 580.8k & 23.2 & 617.3k \\
ESC & Handcrafted & 21.4 & 986.7k & 28.9 & 868.8k & 18.1 & 923.9k & 23.2 & 896.4k \\
Parallel-Probe & Handcrafted & \bf 21.8 & 773.8k & 29.7 & 697.8k & \bf 18.5 & 734.5k & 24.1 & 716.2k \\
AutoTTS ($\beta$=0.5) & Discovered & 19.2 & \bf 283.6k & 28.6 & \bf 250.3k & 14.9 & \bf 257.6k & 21.8 & \bf 254.0k \\
AutoTTS ($\beta$=1.0) & Discovered & 20.9 & 542.2k & \bf 31.1 & 474.7k & 18.0 & 487.1k & \bf 24.6 & 480.9k \\
\midrule
\rowcolor{white}\multicolumn{10}{l}{\textit{Base Model: Qwen3-1.7B}} \\
\midrule
SC @ 64 & Handcrafted & 72.5 & 1025.8k & 44.4 & 1054.1k & 24.2 & 1132.9k & 34.3 & 1093.5k \\
ASC & Handcrafted & 72.3 & 482.6k & 44.4 & 600.9k & 24.2 & 586.3k & 34.3 & 593.6k \\
ESC & Handcrafted & \bf 72.5 & 909.2k & 44.4 & 913.8k & 24.2 & 1014.2k & 34.3 & 964.3k \\
Parallel-Probe & Handcrafted &  68.1 &  748.5k & 44.7 &  775.8k & 22.6 & 860.2k & 33.7  & 818.0k \\
AutoTTS ($\beta$=0.5) & Discovered & 68.5 & \bf 276.3k & 46.7 & \bf 327.9k & 30.5 & \bf 359.1k & 38.6 & \bf 343.5k \\
AutoTTS ($\beta$=1.0) & Discovered & 70.4 & 499.1k & \bf 49.0 & 612.6k & \bf 32.1 & 679.6k & \bf 40.6 & 646.1k \\
\midrule
\rowcolor{white}\multicolumn{10}{l}{\textit{Base Model: Qwen3-4B}} \\
\midrule
SC @ 64 & Handcrafted & 80.0 & 886.8k & 76.6 & 1088.1k & 43.6 & 1168.3k & 60.1 & 1128.2k \\
ASC & Handcrafted & 80.1 & \bf 175.7k & 76.4 & \bf 277.3k & 44.0 & 388.9k & 60.2 & \bf 333.1k \\
ESC & Handcrafted & 80.0 & 528.9k & \bf 76.6 & 793.3k & 43.6 & 990.2k & 60.1 & 891.8k \\
Parallel-Probe & Handcrafted & 79.7 & 688.9k & 76.1 & 806.0k & 44.7 & 872.3k & 60.4 & 839.2k \\
AutoTTS ($\beta$=0.5) & Discovered & 82.0 & 236.7k & 73.8 & 332.3k & 45.7 & \bf 365.0k & 59.8 & 348.7k \\
AutoTTS ($\beta$=1.0) & Discovered & \bf 83.5 & 424.9k & 74.4 & 610.4k & \bf 46.5 & 686.8k & \bf 60.5  & 648.6k \\
\midrule
\rowcolor{white}\multicolumn{10}{l}{\textit{Base Model: Qwen3-8B}} \\
\midrule
SC @ 64 & Handcrafted & 80.4 & 910.8k & 76.7 & 1124.4k & 48.9 & 1267.0k & \bf 62.8 &  1195.7k\\
ASC & Handcrafted & 80.4 & \bf 226.0k & 76.7 & 406.2k & 48.8 & 565.1k & \bf  62.8 & 485.7k \\
ESC & Handcrafted & 80.4 & 459.4k & 76.7 & 793.1k & 48.9 & 1062.1k & \bf  62.8 & 927.6k \\
Parallel-Probe & Handcrafted & 81.5 & 730.8k & \bf 76.9 & 846.7k & 47.1 & 897.2k & 62.0 & 872.0k \\
AutoTTS ($\beta$=0.5) & Discovered & 84.3 & 255.3k & 74.1 & \bf 361.2k & 48.1 & \bf 396.7k & 61.1 & \bf 379.0k  \\
AutoTTS ($\beta$=1.0) & Discovered & \bf 85.8 & 467.4k & 75.8 & 672.4k & \bf 49.5 & 749.1k & 62.7 & 710.8k \\
\bottomrule
\end{tabular}
\end{adjustbox}
\end{table*}

%% file: Fig/scaling_curves.tex
\begin{figure*}[t]
    \centering

    \begin{subfigure}[t]{0.48\textwidth}
        \centering
        \includegraphics[width=\linewidth]{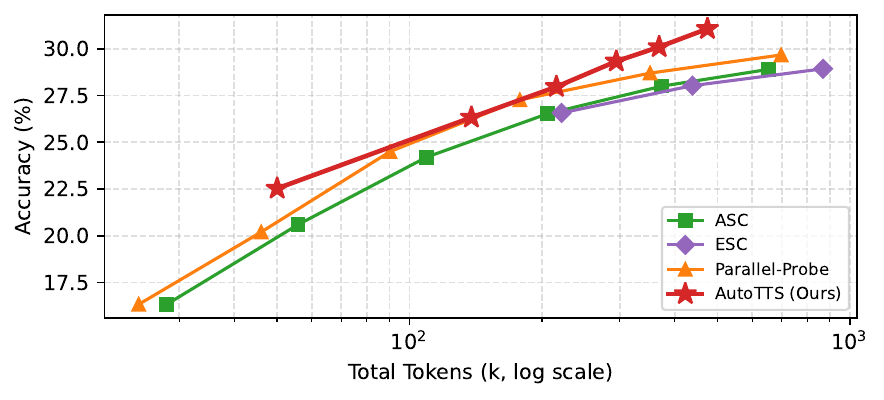}
        \caption{Qwen3-0.6B on AIME25 (held-out)}
        \label{fig:scale_qwen17b_aime25}
    \end{subfigure}
    \hfill
    \begin{subfigure}[t]{0.48\textwidth}
        \centering
        \includegraphics[width=\linewidth]{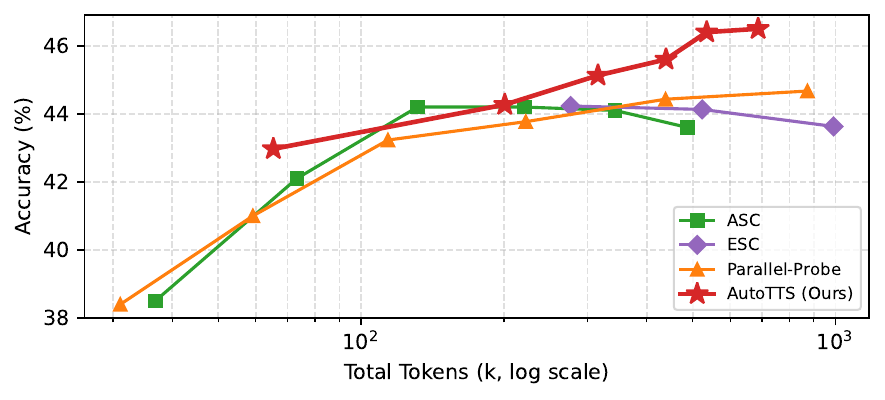}
        \caption{Qwen3-4B on HMMT25 (held-out)}
        \label{fig:scale_qwen4b_hmmt25}
    \end{subfigure}

    \vspace{0.5em}

    \begin{subfigure}[t]{0.48\textwidth}
        \centering
        \includegraphics[width=\linewidth]{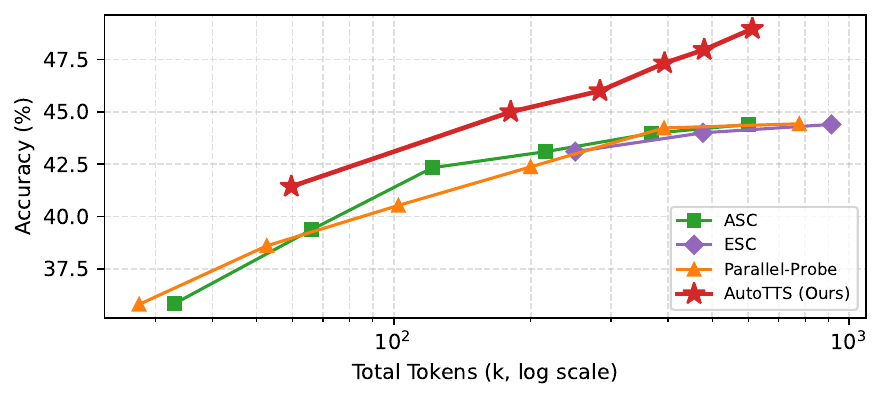}
        \caption{Qwen3-1.7B on AIME25 (held-out)}
        \label{fig:scale_qwen1.7b_aime25}
    \end{subfigure}
    \hfill
    \begin{subfigure}[t]{0.48\textwidth}
        \centering
        \includegraphics[width=\linewidth]{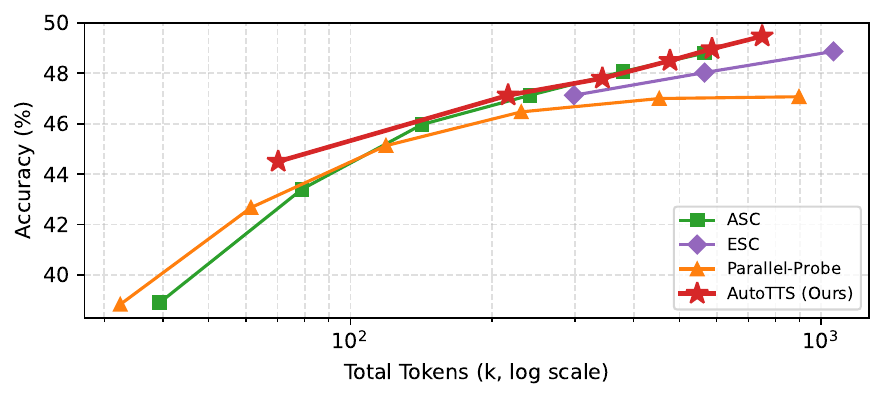}
        \caption{Qwen3-8B on HMMT25 (held-out)}
        \label{fig:scale_qwen8b_hmmt25}
    \end{subfigure}

    \caption{
Accuracy--token scaling curves for discovered and handcrafted controllers.
For handcrafted baselines, scaling is obtained by varying the number of sampled trajectories.
For the discovered controller, scaling is obtained by varying the tradeoff parameter $\beta$.
Across medium-to-large models, the discovered controller traces competitive and often stronger accuracy-efficiency frontiers.
}
    \label{fig:scaling_main}
\end{figure*}

%% file: Raw_Results/generalization.tex
\begin{wraptable}[14]{r}{0.52\textwidth}
\vspace{-2em}
\centering
\small
\setlength{\tabcolsep}{4pt}
\caption{
Generalization beyond the main model and task.
We report accuracy and total inference tokens under each model--dataset setting.
}
\label{tab:generalization_beyond_main}
\begin{tabular}{llcc}
\toprule
\textbf{Method} 
& \textbf{Type}
& \textbf{Acc. $\uparrow$} 
& \textbf{Tokens $\downarrow$} \\
\midrule
\multicolumn{4}{l}{\textit{DeepSeek-R1-Distill-Llama-8B on HMMT25}} \\
SC@64          & Handcrafted & 26.7 & 985.7K \\
ASC            & Handcrafted & 26.5 & 582.7K \\
ESC            & Handcrafted & 26.7 & 952.1K \\
AutoTTS  ($\beta=1$)         & Discovered  & \bf 27.2 & 533.9K \\
AutoTTS  ($\beta=0.5$)         & Discovered  & 26.3 & \bf 279.0K \\
\midrule
\multicolumn{4}{l}{\textit{Qwen3-1.7B on GPQA-Diamond}} \\
SC@64          & Handcrafted & 41.3 & 510.0K \\
ASC            & Handcrafted & 41.0 & 186.3K \\
ESC            & Handcrafted & 41.3 & 391.6K \\
AutoTTS ($\beta=1$)          & Discovered  & \bf 41.6 & 270.1K \\
AutoTTS ($\beta=0.5$)       & Discovered  & \bf 41.6 & \bf 151.0K \\
\bottomrule
\end{tabular}
\vspace{-1.0em}
\end{wraptable}

%% file: Raw_Results/ablation_studies.tex
\begin{table}[t!]
\caption{
Ablation study of the discovery framework. 
We compare the final discovered controller with variants that remove beta parameterization or execution traces. 
For each benchmark, accuracy and total inference tokens are averaged over four models; search cost reports the one-time cost of discovering each strategy.
}
\label{tab:ablation}
\vskip 0.1in
\centering
\small
\setlength{\tabcolsep}{3.5pt}
\begin{adjustbox}{width=\textwidth}
\begin{tabular}{lcc cc cc cc c}
\toprule
\multirow{2}{*}{\textbf{Method}}
& \multicolumn{2}{c}{\textbf{AIME24}}
& \multicolumn{2}{c}{\textbf{AIME25}}
& \multicolumn{2}{c}{\textbf{HMMT25}}
& \multicolumn{2}{c}{\textbf{Avg.}}
& \multirow{2}{*}{\textbf{Search Cost (\$) $\downarrow$}} \\
\cmidrule(lr){2-3}
\cmidrule(lr){4-5}
\cmidrule(lr){6-7}
\cmidrule(lr){8-9}
& \textbf{Acc. $\uparrow$}
& \textbf{Tokens $\downarrow$}
& \textbf{Acc. $\uparrow$}
& \textbf{Tokens $\downarrow$}
& \textbf{Acc. $\uparrow$}
& \textbf{Tokens $\downarrow$}
& \textbf{Acc. $\uparrow$}
& \textbf{Tokens $\downarrow$}
& \\

\midrule
\rowcolor{gray!8}
Ours
& \textbf{65.2} & 483.4K
& \textbf{57.6} & 592.5K
& \textbf{36.5} & 650.7K
& \textbf{53.1} & 575.5K
& 39.9 \\

\midrule
\multicolumn{10}{l}{\textit{Ablation on beta-controlled search space}} \\

w/o Beta Parameterization
& 60.7 & \bf 81.2K
& 54.4 & \bf 93.9K
& 31.9 & \bf 104.7K
& 49.0 & \bf 93.3K
& 46.4 \\

\midrule
\multicolumn{10}{l}{\textit{Ablation on history design}} \\

w/o Execution Traces
& 64.0 & 703.1K
& 56.7 & 823.7K
& 34.1 & 946.1K
& 51.6 & 824.3K
& \bf 30.9 \\

\bottomrule
\end{tabular}
\end{adjustbox}
\end{table}

%% file: Raw_Results/cost_efficiency.tex

%% file: Fig/self_evolving.tex
\begin{figure*}
    \centering
    \includegraphics[width=\linewidth]{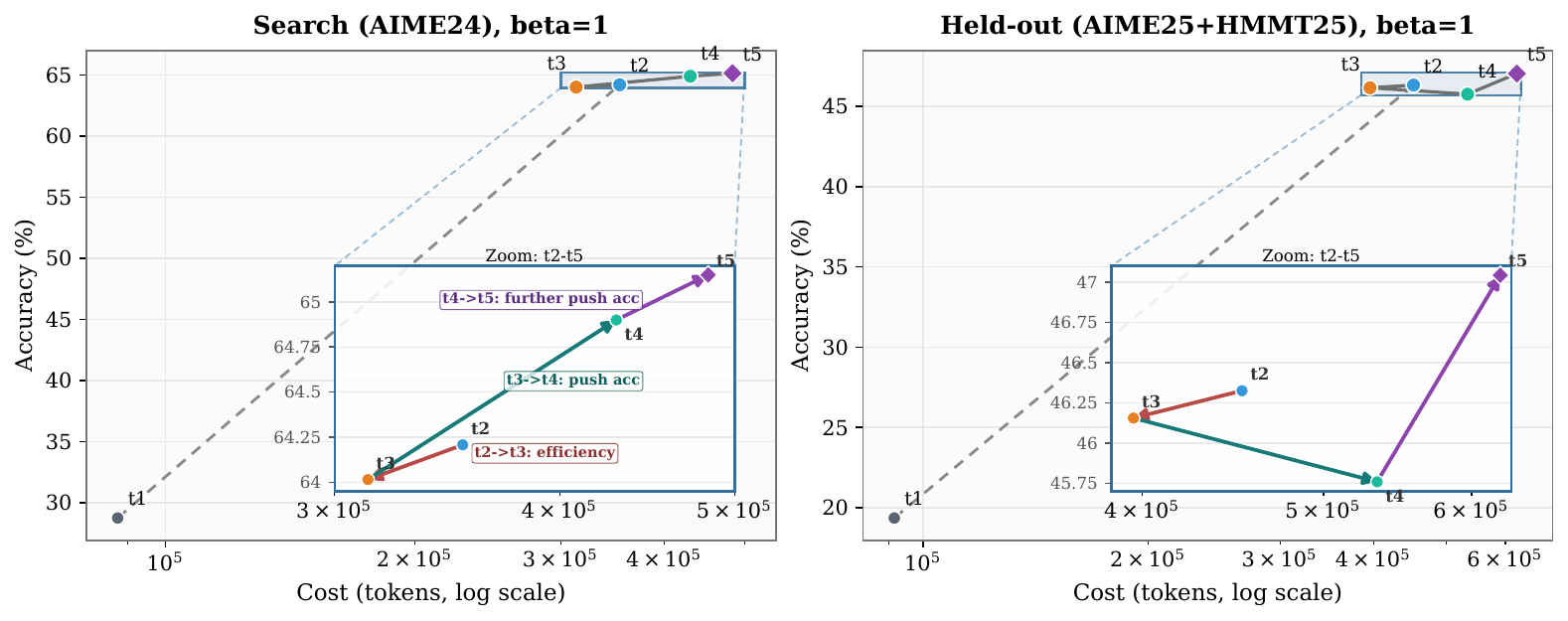}
    \caption{
The trajectory illustrates how the proposer iteratively corrects the accuracy--cost trade-off.
The initially proposed controller over-optimizes for token efficiency, which leads to a substantial accuracy drop.
After observing this degradation, the proposer increases the budget in the next turn to recover performance.
Subsequent turns continue this feedback loop, alternating between efficiency-oriented adjustments and performance recovery, gradually moving toward a better accuracy--cost allocation.
}
    \label{fig:self_evolving}
\end{figure*}

%% file: Main/related_work.tex
\section{Related Work}
\paragraph{Efficient Parallel Reasoning}
To alleviate the high computational overhead associated with fixed-budget search strategies, recent studies have turned to dynamic resource management. For instance, Aggarwal et al. \cite{aggarwal2023let} and Li et al. \cite{li2024escape} introduce adaptive stopping criteria triggered by consensus, whereas Wang et al. \cite{wang2025make} tailor the number of allocated samples to the complexity of the query. Aside from simply reducing the sample count, confidence-guided strategies assign weights to different reasoning trajectories, enabling the discovery of optimal solutions with minimal samples \cite{huang2025efficient,Taubenfeld2025ConfidenceIS,Fu2025DeepTW}. Nevertheless, these techniques largely rely on sequential sampling processes, which bottlenecks the hardware utilization necessary for efficient parallel execution. In a more granular approach, recent works such as Dynamic Self-Consistency \cite{wan2025reasoning}, Self-Truncation \cite{wang2025sampling}, DeepPrune \cite{tu2025deepprune}, Step \cite{liang2026hidden}, and Slim-SC \cite{hong2025slim} intervene during generation to discard unpromising branches, thereby preventing redundant calculations on flawed paths. 

\paragraph{Efficient Sequential Reasoning}
To enhance reasoning depth without the need for further training, current literature frequently employs dynamic early-exit strategies. One prominent approach relies on tracking uncertainty indicators: Wang et al. \cite{wang2025entropy} and Sharma et al. \cite{sharma2025think} use post-reasoning or sequence-level entropy to gauge confidence, while Yong et al. \cite{yong2025think} achieve this through empirical estimation using beam search or repeated rollouts. Another line of work bases halting decisions on output consistency, where the convergence of intermediate answers serves as an indicator to terminate generation \cite{liu2025answer,Mao2025EarlySC,fu2025reasoning,zhang2025alphaone}. Moving beyond surface-level output metrics, Zhang et al. \cite{zhang2025reasoning} advocate for internal self-verification by probing hidden states, which empowers models to cease inference upon reaching a predefined confidence threshold \cite{Yang2025DynamicEE}. 

\paragraph{From AutoML to Agentic Discovery.} Algorithm discovery has evolved from classical AutoML~\citep{zoph2016neural, elsken2019neural} to LLM-driven program search, where FunSearch~\citep{romera2024mathematical}, EoH~\citep{liu2024evolution}, AlphaEvolve~\citep{novikov2025alphaevolve}, and ADAS~\citep{hu2024automated} use LLMs to iteratively propose and refine algorithms in code. Meta-Harness~\citep{lee2026meta} further advances this paradigm by exposing full execution histories to the proposer, enabling targeted diagnosis of failure modes rather than relying on scalar feedback alone. TTT-Discover~\citep{yuksekgonul2026learning} and ThetaEvolve~\citep{wang2025thetaevolve} push further by updating model weights via RL at test time. While these methods demonstrate the power of automatic discovery in various domains, \textbf{TTS strategy design has remained hand-crafted}—no prior work has formulated it as an algorithmic search problem. AutoTTS proposes this reframing: by casting TTS as controller synthesis over a structured control space, TTS research can be advanced through environment-driven discovery. The key to making this search affordable is instantiating it over a replay MDP built from pre-collected trajectories, where controller evaluation requires no additional LLM calls—directly addressing the evaluation cost bottleneck that limits prior discovery methods.

%% file: Main/Conclusion.tex
\section{Conclusion}

In this work, we introduce AutoTTS, an environment-driven framework 
that reframes test-time scaling strategy design as an automated 
discovery problem. By shifting the human role from hand-crafting 
individual branching, pruning, and stopping heuristics to constructing 
replayable discovery environments, AutoTTS enables an explorer LLM to 
automatically synthesize controllers that outperform strong handcrafted 
baselines across model scales and benchmarks. At the core of this 
approach is the insight that effective TTS strategies need not be 
manually engineered---they can be discovered automatically given the 
right environment. The one-time discovery cost of \$39.87 and 160 
minutes demonstrates that environment-driven discovery is practical 
today, paving the way for a new paradigm where TTS research advances 
through environment design rather than strategy design.

%% file: Main/appendix.tex
\newpage
\appendix

\section{Limitation}
\label{app:limitation}
AutoTTS demonstrates that effective TTS strategies can be automatically discovered through environment-driven search at minimal cost, with the discovered controller generalizing across model scales and benchmarks. The current instantiation constructs environments for width--depth TTS 
control. It would be interesting to extend the action set and build richer environments that support more complex control structures. Additionally, the discovery process currently relies on a frontier coding agent; exploring whether open-source coding agents can achieve comparable discovery performance is an interesting direction for future work.

\section{Broader Impact}
\label{app:broader_impact}
AutoTTS introduces an environment-driven framework that 
automatically discovers effective test-time scaling strategies, shifting 
the human role from manually designing one-off algorithms to constructing 
reusable discovery environments. This has positive societal impact by 
improving the efficiency of LLM inference at test time, reducing 
computational cost and making capable reasoning more accessible. More 
broadly, this work proposes a paradigm shift in how human expertise is 
invested: rather than encoding domain understanding into individual 
algorithms, researchers encode it into reusable environments that can 
be repeatedly leveraged by AI agents, allowing human insight to compound 
over time and significantly accelerating progress in TTS research and 
beyond. As foundational research, we do not foresee direct negative 
societal impacts, though more efficient reasoning systems could in 
principle be misused to accelerate harmful applications, a risk shared 
broadly across inference optimization research.

\section{Discovery Agent Prompt}
\label{app:prompt}

The following prompt is provided to the explorer LLM (Claude Code) at the start of each discovery round. It defines the environment interface, design constraints, and search objectives for the \texttt{OptimalController}.

\begin{tcolorbox}[
    colback=gray!5,
    colframe=gray!50,
    title=Discovery Agent Prompt,
    fonttitle=\bfseries,
    breakable
]
\begin{Verbatim}
You are a proposer for training-free reasoning controller discovery in a 2D probing environment.

Your goal is not to solve the reasoning task directly.
Your goal is to discover and implement a reusable controller algorithm over offline branch-probing traces.

Reasoning-time control is formulated as a 2D optimization problem:
- width = how many branches to explore
- depth = how far each active branch has been probed

For each question, offline data provides a set of branches.
Each branch contains:
- a sequence of intermediate probe answers
- a total branch token budget
- a final answer

Because branch lengths differ, the 2D probing state is irregular rather than a perfectly aligned matrix.

Environment API (on the `question` object passed to `solve`):
- `question.probe_new()` -> `(answer, branch_index, is_finish)`
  starts a new reasoning branch, lets it decode for a fixed token chunk, and returns that branch's current answer, its stable `branch_index`, and whether the branch has already finished in this call.
- `question.probe_more(branch_index)` -> `(answer, is_finish)`
  continues an already activated branch for another fixed token chunk, and returns its updated answer plus whether the branch is now finished.
- `question.get_new_branch_final_answer()` -> `answer`
  consumes one whole fresh branch at full token cost and returns only its final answer (no intermediate probing). Use this for full-read style baselines such as majority voting / ASC / ESC.
- `question.get_seq_and_total_tokens()` -> `(seq_tokens, total_tokens)`
  returns cumulative cost accounting for the current question. Use for diagnostics only; do not rely on it as an oracle stop signal.

Helper utilities available in `code_base/method.py` (prefer reusing them, not re-implementing):
- `_safe_probe_new(question)`, `_safe_probe_more(question, index)`, `_safe_full_read(question)`
  swallow `IndexError`/`ValueError` when a branch is exhausted and return `None` instead of raising.
- `_majority_answer(answers)`, `_vote_stats(answers)`, `_beta_majority_confidence(top1, top2)`, `_weighted_vote(pairs)`
  standard voting / confidence primitives already used by ASC and Parallel-Probe.

Important:
- Both `probe_new()` and `probe_more(index)` represent actual reasoning progress, not just answer inspection.
- Each action advances decoding by a fixed chunk of tokens before exposing the branch's current answer.
- Early probed answers are often unreliable: shallow branch predictions may be incorrect, unstable, or misleading, and only become accurate after sufficient reasoning depth.
- `get_new_branch_final_answer()` spends the branch's full budget in one call, so mixing it with `probe_new/probe_more` inside the same controller changes cost accounting semantics; pick one style per controller unless the mixing is intentional.

Design the controller around these observable primitive operations only.
Do not use hidden future information, oracle signals, dataset-specific shortcuts, or hardcoded answers.

Objectives:
1. maximize answer accuracy
2. minimize total token cost

Accuracy is the primary objective.
Among controllers with similar accuracy, prefer lower total cost.
The goal is to improve held-out accuracy-efficiency behavior, not merely search-set numbers.

The target is a reusable controller algorithm that maps the current 2D probing state to decisions such as:
- whether to start a new branch
- which active branch or branches to probe next
- whether to continue widening exploration
- whether to stop probing
- how to aggregate answers at termination

Implementation constraints:
- Directly edit `OptimalController` in `code_base/method.py`. `code_base/method.py` is reset from `code_base/method.template.py` at the start of every round, so your patch must be self-contained within `method.py` and must not rely on out-of-file state.
- Keep the implementation executable by `code_base/eval.py` without interface or runtime errors. In particular:
  - subclass `LLMDesignedMethod`, keep `NAME`, `__init__(self, config: Optional[Dict[str, Any]] = None)`, and `solve(self, question) -> Optional[str]` compatible with the existing call site in `data_loader.ModelandTask.evaluate`.
  - accept tunable arguments either as explicit kwargs or via the `config` dict; do not change the constructor signature that `eval.py` uses (currently `OptimalController(config={{...}})`).
- Reuse shared helpers and `ASCMethod` / `ESCMethod` / `Parallel_Probe` infrastructure that already sits in `method.py` whenever it is natural; do not duplicate or rewrite those seed classes.
- Keep the patch focused; avoid broad refactoring. Do not modify shared helpers, the seed methods, the evaluator, data loader, datasets, or any framework files.
- Read other files only for context.

Tracing expectation (align with the existing seeds):
- All seed methods in `method.py` expose per-question traces via `MethodTraceRecorder` with a small, consistent surface: `self.trace_recorder`, `_reset_trace()`, `_trace_step(event, goal, step_input, step_output, state, decision)`, `get_last_trace()`, and `solve_with_trace(question)`.
- `data_loader.ModelandTask.evaluate` detects `hasattr(function, "get_last_trace")` and writes the recorded trace to the per-round `training_results/matrix_results_<Model>/<dataset>_trace_new_api.jsonl`. Your `OptimalController` should implement the same surface so its behavior is diagnosable alongside the seeds.
- Use `_reset_trace()` at the top of `solve`, emit a `start` event after reset, one event per decision point (e.g., `init_branches`, `forward`, `update_states`, `prune`, `terminate_check`, `finish`), and make sure a single terminal `finish` event is recorded on every return path.
- Keep per-step trace payloads small and JSON-serializable (primitive types, short lists); avoid dumping full branch histories at every step.

Fair comparison and budget requirement:
- The controller should support a maximum branch budget of `max_branch = 64`, matching the strong seed baselines whenever applicable (e.g., `ASCMethod(max_samples=64)`, `ESCMethod(max_samples=64, ...)`, `Parallel_Probe(num_chains=64, ...)`).
- Treat `max_branch = 64` as an available budget ceiling, not as a fixed branch count that must always be used, and not as a smaller artificial cap used to win by restriction.
- A meaningful improvement means using the same budget ceiling more effectively through adaptive control.

Adaptive control requirement:
- The controller should be genuinely adaptive under the shared `max_branch = 64` ceiling.
- It should adaptively allocate budget across both width and depth based on the current 2D probing state.
- It should decide when to widen exploration, when to deepen selected branches, when to stop widening, and when to terminate altogether.
- Avoid mostly fixed probing schedules or nearly fixed expansion-and-stop templates.
- A controller that adaptively widens but then uniformly deepens all active branches is not fully sufficient; depth allocation should also be adaptive whenever appropriate.

Budget-family requirement:
- The controller may expose a single high-level budget parameter, e.g. `beta`, that controls budget aggressiveness.
- `beta` should not hardcode an exact token target for each question.
- Instead, `beta` should induce a monotonic shift from conservative to aggressive budget use while the controller still adaptively allocates budget based on the current 2D probing state.
- `beta` should serve as the primary knob for tracing an accuracy-cost frontier under the shared `max_branch = 64` ceiling.
- Accept `beta` through `config={{"beta": ...}}` so that `eval.py` sweeps like `for beta in betas: OptimalController(config={{"beta": beta}})` work unchanged.

Conservative-anchor requirement for `beta`:
- The `beta` scale should have a clear conservative anchor.
- In particular, when `beta = 1`, the controller should behave conservatively rather than aggressively:
  - it should be reluctant to terminate based on shallow or newly formed consensus,
  - it should avoid overly aggressive early pruning,
  - it should prefer gathering reasonably mature evidence before making irreversible stop/prune decisions,
  - and it should be allowed to approach near-full use of the shared budget ceiling when the state remains ambiguous.
- Lower-budget behavior for smaller effective spending should come mainly from more selective width/depth allocation, not from brittle premature stopping rules.

Monotonicity requirement:
- As `beta` increases, the controller should become systematically more willing to widen exploration, deepen uncertain branches, defer termination, and in general spend more of the available budget.
- As `beta` decreases, the controller may become more conservative in spending, but should not rely on overly brittle shallow-consensus stopping or aggressive early exits that noticeably damage accuracy.

Coverage requirement:
- By varying `beta`, it should be possible to trace a meaningful frontier from low-budget regimes up to near-full use of the shared `max_branch = 64` budget ceiling.
- In the high-`beta` regime, the controller should be capable of behaving close to a full-budget, accuracy-first policy rather than remaining artificially frugal.
- Prefer a coherent budget-controlled controller family over many loosely related controller variants.

Single-knob schedule requirement:
- All behavior-critical hyperparameters of the controller (e.g. warm_up, prune_patience, stability window `T`, confidence threshold, window size, any per-branch budget cap, per-step probe burst, termination thresholds) must be deterministic functions of `beta` computed inside `OptimalController`. Given a single `beta` value, the controller must be fully determined — no additional tunable knobs are exposed to `eval.py`.
- `config` should only carry `beta` (and at most a few fixed structural constants such as the `max_branch = 64` ceiling that are not tuned between runs). Do not accept multiple independent hyperparameter fields alongside `beta`.
- Implement the mapping as a named helper such as `def _schedule(self, beta) -> dict` that returns every internal hyperparameter value as a function of `beta`, and call it once in `__init__`. This keeps the schedule explicit, reviewable, and easy to replace.
- The schedule should be smooth and monotonic with `beta` wherever the Monotonicity requirement applies (e.g. warm_up, T, patience, per-branch cap should be non-decreasing in `beta`; early-stop thresholds should be non-increasing in aggressiveness as `beta` grows). Avoid piecewise logic with many hand-placed breakpoints; prefer a small number of simple analytic forms (linear, power, sigmoid, clipped ramps) with a handful of fixed structural constants.
- You are encouraged to tune the `beta -> hyperparameter` schedule design with a local sandbox: prototype alternative schedule forms, plot the induced accuracy-cost frontier across the evaluated `beta` grid against the seed baselines, and keep the simplest schedule that achieves a robust frontier. Do not leave the sandbox-only code in `OptimalController`; the committed class should contain only the final schedule plus the controller logic.
- Any design-time fitting of the schedule must use only material already allowed by this prompt (prior proposals, seed traces, the offline probing environment). Do not peek at gold answers or held-out splits.

History-aware proposing:
- Seed baseline source code lives directly in `code_base/method.py` (and the reset template `code_base/method.template.py`):
  `ASCMethod` (sequential full-read + Beta-majority confidence early stop),
  `ESCMethod` (windowed full-read + unanimity early stop),
  `Parallel_Probe` (parallel chains with warm-up, off-track pruning, stable-majority termination).
- Reference benchmarks and per-step traces for the seeds are at `code_base/history/seed_algorithms/matrix_results_<Model>/<dataset>_raw_new_api.csv` and `code_base/history/seed_algorithms/matrix_results_<Model>/<dataset>_trace_new_api.jsonl`. Inspect them to learn the current accuracy-cost frontier and the shape of per-question traces for each baseline family.
- Prior proposals from previous rounds are archived at `code_base/history/rNNNN_<timestamp>_<uid>/`, each containing:
  - `method.py` — the exact `OptimalController` snapshot that was evaluated that round,
  - `proposal_results/matrix_results_<Model>/*.csv` — per-method accuracy/cost numbers,
  - `proposal_results/matrix_results_<Model>/*_trace_new_api.jsonl` — per-question traces,
  - plus any manifest files.
  The round index grows over time; read the latest few rounds first to understand what has just been tried.
- Before proposing a new controller, inspect both the prior proposals (with their traces) and the seed baselines. Use them to understand what has already been tried, what failed, what currently works well, and what the current performance frontier looks like.

Novelty requirement:
- The proposal must be meaningfully different from the seed algorithms, especially `ASCMethod`, `ESCMethod`, and `Parallel_Probe`.
- Do not simply reproduce, lightly reweight, or trivially combine existing algorithms.
- Small threshold tuning, slight stage reordering, or cosmetic variants do not count as sufficient novelty.
- If building on prior ideas, introduce a clear new mechanism that substantively changes the controller's decision logic.
- However, novelty is not the goal by itself: prefer a novel mechanism with a plausible path to robust held-out performance, not a complicated but brittle idea.

Robustness and parameter discipline:
- Fewer hyperparameters is strictly better. Each additional tunable parameter adds a degree of freedom that can be implicitly overfit to the search set and degrade held-out robustness, so the default stance is to introduce as few parameters as possible and to delete any parameter that does not demonstrably change behavior.
- Prefer controller families with few, interpretable, behavior-critical parameters.
- Keep the number of independently tuned behavior-critical parameters as small as possible.
- Prefer a controller family governed by one high-level budget parameter (e.g. `beta`) plus at most a few fixed structural constants.
- Avoid many tightly coupled thresholds, patience counters, or brittle stage-specific hyperparameters unless clearly necessary.
- Do not cherry-pick a fragile best operating point.
- Prefer stable operating regions and smooth budget-performance curves over sharp search-set optima.
- Do not expose tunable knobs other than `beta`. Every behavior-critical hyperparameter must be a function of `beta` (see the Single-knob schedule requirement); if you believe an additional externally tunable parameter is unavoidable, explicitly justify why it cannot be folded into the `beta` schedule before introducing it.
- Small gains that require many sensitive parameters are not desirable; a slightly weaker controller with fewer hyperparameters is preferred over a slightly stronger one that needs careful multi-knob tuning.

Performance requirement:
- The proposed controller should aim to outperform the seed algorithms in the overall accuracy-efficiency tradeoff.
- Desirable outcomes include:
  - similar answer accuracy with fewer total tokens, or
  - better answer accuracy under a similar token budget.
- Do not sacrifice too much accuracy for small token savings.
- The goal is a novel controller that is competitive with strong seed algorithms and plausibly stronger on held-out accuracy-efficiency behavior.

Do not redesign the evaluation framework.

Required workflow:
1. Inspect the current codebase and understand the environment abstraction (`code_base/data_loader.py`, `code_base/eval.py`).
2. Inspect `code_base/method.py` (and `code_base/method.template.py`) to read the seed implementations `ASCMethod`, `ESCMethod`, `Parallel_Probe`, the shared helpers, and the `OptimalController` stub you must fill in.
3. Inspect `code_base/history/seed_algorithms/matrix_results_<Model>/` for the seed accuracy-cost frontier and per-step traces.
4. Inspect the latest few `code_base/history/rNNNN_<ts>_<uid>/` folders for prior `OptimalController` proposals (`method.py` snapshot) and their `proposal_results/` (CSV + trace JSONL).
5. Summarize the core ideas of prior proposals and seed algorithms.
6. Identify what has already been tried, what failed, and what would count as genuinely different.
7. Propose a new controller that is novel, adaptive, and plausibly stronger in held-out accuracy-efficiency behavior.
8. Prefer a coherent budget-controlled controller family, ideally governed by `beta`, over many independent thresholds.
9. Implement the controller by directly editing `OptimalController` in `code_base/method.py`. Wire the `MethodTraceRecorder` / `_trace_step` / `solve_with_trace` surface in the same style as the seeds so traces are emitted into `training_results/matrix_results_<Model>/<dataset>_trace_new_api.jsonl`.
10. Ensure constructor, signature, inheritance/interface behavior, and return behavior remain compatible with `code_base/eval.py` (which currently instantiates `OptimalController(config={{"beta": beta}})`).
11. Keep the patch focused and avoid unnecessary changes.
12. Summarize:
   - the controller idea
   - whether history was used (which rounds and which seed traces)
   - what lessons were extracted from prior proposals and seed algorithms
   - what makes this controller genuinely different from `ASCMethod`, `ESCMethod`, and `Parallel_Probe`
   - why this controller is adaptive in width-depth budget allocation
   - how it uses the shared `max_branch = 64` budget ceiling
   - whether and how `beta` is used to provide full budget coverage from low budget to near-full budget
   - the exact `beta -> hyperparameter` schedule (analytic form per hyperparameter, fixed structural constants, monotonicity direction) and how it was tuned in the sandbox (what alternatives were tried, why the committed form was selected)
   - how parameter complexity was controlled (in particular, confirmation that no tunable knob other than `beta` is exposed)
   - how tracing was instrumented (events emitted, terminal `finish` coverage, payload size discipline)
   - why this proposal may improve the held-out accuracy / total-cost tradeoff
   - possible risks or failure modes
\end{Verbatim}
\end{tcolorbox}

\section{Discovered TTS Program}
\label{app:controller}

The discovered controller, which we term the \textit{Confidence Momentum 
Controller} (CMC), reveals four non-obvious mechanisms.

\textbf{Trend-based stopping.}
Rather than gating termination on instantaneous confidence, CMC maintains 
an EMA of pool confidence and stops only when both the EMA level is high 
and the trend is non-negative, preventing premature stopping on transient 
confidence spikes.

\textbf{Coupled width--depth control.}
Widening and deepening are linked through the EMA delta: strong confidence 
gains suppress new branch spawning, while stagnation or regression triggers 
widening, creating a closed feedback loop absent in all hand-crafted baselines.

\textbf{Alignment-aware depth allocation.}
Each round, branches whose latest answer matches the pool winner receive 
\texttt{burst\_aligned} probe steps. This concentrates computation on 
the emerging consensus while still advancing all active branches.

\textbf{Conservative branch abandonment.}
A branch is only abandoned after persistently deviating for 
\texttt{abandon\_patience} consecutive rounds, with at least two active 
branches always preserved.

Together, these mechanisms represent a level of coordinated complexity 
that would be difficult to arrive at through manual intuition alone.

\begin{tcolorbox}[
    colback=gray!5,
    colframe=gray!50,
    title=Discovered Controller (OptimalController),
    fonttitle=\bfseries,
    breakable
]
\begin{Verbatim}
class OptimalController(LLMDesignedMethod):
    """
    Confidence Momentum Controller (CMC).

    Core idea
    ---------
    All prior proposals (IBC, SCR, DGCC) share the same fundamental stopping
    signal: "instantaneous" Beta-majority confidence computed from the
    completed-answer pool at the current step.  This is susceptible to
    single-step confidence spikes: a lucky early cluster of identical answers
    can fire the gate prematurely before the distribution has stabilised.

    CMC replaces the instantaneous confidence gate with a **momentum-aware**
    gate:
      - Track an exponential moving average (EMA) of pool confidence over
        the last `T_ema` rounds: ema_conf = alpha * conf + (1 - alpha) * ema_conf
      - Track the recent improvement delta: delta = ema_conf - ema_conf_prev
      - Gate fires when BOTH of the following hold:
          (a) ema_conf >= conf_thresh  (level requirement)
          (b) delta >= -slack          (non-deteriorating momentum; slack is
              a small tolerance that prevents stopping on a declining signal)
      This means the controller cannot stop on a one-round spike; the EMA
      must be high and not actively falling.

    Adaptive depth allocation via probe-age priority
    ------------------------------------------------
    Each active unfinished branch tracks `probe_count` (how many probe steps
    it has received).  In each round the controller allocates a per-round
    probe budget of `probe_budget` steps distributed across active branches
    using a **priority queue** sorted by probe_count descending.  The most-
    invested branches get served first (up to `burst_senior` extra steps
    each), then remaining budget goes to less-invested branches.
    This concentrates depth on branches that are closest to completion while
    still advancing younger branches, rather than uniform or purely aligned-
    biased allocation (SCR) or lazy sleeping (DGCC).

    Three-tier branch classification
    ---------------------------------
    After warm_up:
      - "aligned":  latest answer == pool_winner
      - "deviant":  latest answer != pool_winner, disagreed for >= 1 round
      - "neutral":  no pool winner yet, or first round of disagreement
    Tier affects the per-branch probe multiplier:
      aligned  -> multiplier = `burst_aligned`  (e.g. 2 at high beta)
      neutral  -> multiplier = 1
      deviant  -> multiplier = 1, but if deviant for >= `abandon_patience`
                  rounds the branch is abandoned

    Confidence-trend widening
    -------------------------
    Widening (spawning new branches) is driven by whether the confidence
    *trend* (delta) is positive and large, or weak/negative:
      - if delta > trend_thresh: confidence is accelerating -> no widening
        (we're on track to stop soon)
      - if delta <= trend_thresh: plateau or regression -> widen by
        `widen_burst` new branches, up to max_branch ceiling
    This directly couples width decision to whether deepening is yielding
    evidence-quality gains, a feedback loop not present in prior proposals.

    Beta schedule
    -------------
    All hyperparameters are deterministic functions of a single beta in [0,1].
    beta=0 -> conservative (few branches, low EMA inertia, easier to stop)
    beta=1 -> near-full budget (many branches, high inertia, harder to stop)

    Novelty vs prior work
    ---------------------
    ASC / ESC: full reads; no incremental probing.
    Parallel_Probe: fixed cohort; instantaneous majority; no pool/completion
      distinction; no EMA.
    IBC (r0001): instantaneous pool confidence gate; uniform 1-step probing;
      1-branch-per-round widening; no EMA or trend.
    SCR (r0002): asymmetric burst (aligned gets more steps); plateau-triggered
      widening; instantaneous gate; no EMA.
    DGCC (r0003): dual instantaneous gate (primary + soft corroboration);
      lazy sleeping for locked branches; vote-gap proportional widening;
      no EMA momentum.
    CMC: replaces ALL instantaneous gates with a single EMA momentum gate;
      introduces probe-age priority scheduling (neither uniform nor burst-
      aligned-only); confidence-trend widening (neither plateau nor vote-gap);
      three-tier classification is a natural simplification vs DGCC's dual
      gate without adding extra hyperparameters.
    """

    NAME = "optimal_controller"

    # Fixed structural constants
    _MAX_BRANCH   = 64
    _MAX_OUTER    = 500   # hard cap on outer loop iterations

    # ------------------------------------------------------------------
    # Schedule: single beta knob -> all hyperparameters
    # ------------------------------------------------------------------

    def _schedule(self, beta: float) -> dict:
        """
        All schedules are smooth analytic functions of beta in [0,1].
        Monotonicity:
          - Parameters controlling budget use (n_init, max_branch_use,
            burst_aligned, widen_burst, warm_up, abandon_patience, T_ema)
            are NON-DECREASING in beta.
          - conf_thresh is NON-DECREASING in beta (harder to stop -> more budget).
          - trend_thresh is NON-INCREASING in beta (easier to trigger widening
            at high beta -> more budget via wider exploration).
          - ema_alpha is NON-DECREASING in beta (more inertia at high beta;
            EMA reacts more slowly -> harder to stop on one spike).

        Forms (all clipped to [0,1] after computation where applicable):

          n_init         = max(2, round(2 + 6*b))         [2,  8]
          max_branch_use = min(64, round(4 + 60*b))       [4, 64]
          warm_up        = max(2, round(2 + 8*b))         [2, 10]
          abandon_patience = max(3, round(3 + 9*b))       [3, 12]

          T_ema   = max(2, round(2 + 6*b))                [2,  8]
            Window length for EMA; longer window = more inertia.
          ema_alpha = 0.3 + 0.5 * b                       [0.30, 0.80]
            EMA blending factor (higher = more weight on current value;
            note: convention here is ema = alpha*new + (1-alpha)*old so
            higher alpha gives LESS inertia; but we want more budget at
            high beta meaning more inertia, so use 1 - alpha in update:
            Actually: ema = (1-alpha)*ema + alpha*conf.  Higher alpha ->
            faster EMA convergence (less inertia).  We want MORE inertia at
            high beta so alpha should be LOWER at high beta.
            Schedule: ema_alpha = 0.70 - 0.40 * b        [0.30, 0.70]
            (NON-INCREASING: lower alpha = slower EMA = more inertia = more budget)

          conf_thresh = 0.85 + 0.12 * b                   [0.85, 0.97]
            (NON-DECREASING: harder to stop at high beta)

          delta_slack = 0.04 - 0.03 * b                   [0.01, 0.04]
            Tolerance for declining EMA that still allows stopping.
            NON-INCREASING in beta (stricter at high beta -> harder to stop)

          burst_aligned = max(1, round(1 + 2*b))          [1, 3]
            Extra probe steps for aligned branches per round.

          widen_burst = max(1, round(1 + 3*b))            [1,  4]
            Number of new branches to spawn per widening event.

          trend_thresh = 0.04 - 0.03 * b                  [0.01, 0.04]
            Maximum EMA delta below which widening is triggered.
            NON-INCREASING: at high beta, widening triggers even when
            confidence is growing moderately (more exploration).

          min_complete = max(2, round(2 + 3*b))           [2,  5]
            Minimum completed branches before any gate is eligible.
        """
        b = max(0.0, min(1.0, float(beta)))

        n_init           = max(2, round(2  + 6  * b))
        max_branch_use   = min(self._MAX_BRANCH, round(4 + 60 * b))
        warm_up          = max(2, round(2  + 8  * b))
        abandon_patience = max(3, round(3  + 9  * b))

        # EMA parameters
        T_ema            = max(2, round(2  + 6  * b))
        ema_alpha        = 0.70 - 0.40 * b          # [0.30, 0.70], NON-INCREASING

        # Stopping gate
        conf_thresh      = 0.85 + 0.12 * b          # [0.85, 0.97]
        delta_slack      = 0.04 - 0.03 * b          # [0.01, 0.04]

        # Depth allocation
        burst_aligned    = max(1, round(1 + 2 * b))  # [1, 3]

        # Widening
        widen_burst      = max(1, round(1 + 3 * b))  # [1, 4]
        trend_thresh     = 0.04 - 0.03 * b           # [0.01, 0.04], NON-INCREASING

        # Gate eligibility
        min_complete     = max(2, round(2 + 3 * b))  # [2, 5]

        return {
            "n_init":           n_init,
            "max_branch_use":   max_branch_use,
            "warm_up":          warm_up,
            "abandon_patience": abandon_patience,
            "T_ema":            T_ema,
            "ema_alpha":        round(ema_alpha, 4),
            "conf_thresh":      round(conf_thresh, 4),
            "delta_slack":      round(delta_slack, 4),
            "burst_aligned":    burst_aligned,
            "widen_burst":      widen_burst,
            "trend_thresh":     round(trend_thresh, 4),
            "min_complete":     min_complete,
        }

    # ------------------------------------------------------------------
    # Init
    # ------------------------------------------------------------------

    def __init__(self, config: Optional[Dict[str, Any]] = None):
        super().__init__(config)
        self._beta           = float((config or {}).get("beta", 0.5))
        sched                = self._schedule(self._beta)
        self.n_init          = sched["n_init"]
        self.max_branch_use  = sched["max_branch_use"]
        self.warm_up         = sched["warm_up"]
        self.abandon_patience = sched["abandon_patience"]
        self.T_ema           = sched["T_ema"]
        self.ema_alpha       = sched["ema_alpha"]
        self.conf_thresh     = sched["conf_thresh"]
        self.delta_slack     = sched["delta_slack"]
        self.burst_aligned   = sched["burst_aligned"]
        self.widen_burst     = sched["widen_burst"]
        self.trend_thresh    = sched["trend_thresh"]
        self.min_complete    = sched["min_complete"]
        self.trace_recorder  = MethodTraceRecorder()

    # ------------------------------------------------------------------
    # Trace surface (matches seed style)
    # ------------------------------------------------------------------

    def _reset_trace(self) -> None:
        self.trace_recorder = MethodTraceRecorder()

    def _trace_step(
        self,
        *,
        event: str,
        goal: str,
        step_input: Dict[str, Any],
        step_output: Any,
        state: Dict[str, Any],
        decision: str,
    ) -> None:
        self.trace_recorder.add_step(
            event=event,
            goal=goal,
            input=step_input,
            output=step_output,
            state=state,
            decision=decision,
        )

    def get_last_trace(self) -> List[Dict[str, Any]]:
        return self.trace_recorder.to_list()

    def solve_with_trace(self, question) -> Dict[str, Any]:
        answer = self.solve(question)
        return {"answer": answer, "trace": self.get_last_trace()}

    # ------------------------------------------------------------------
    # Internal helpers
    # ------------------------------------------------------------------

    def _pool_stats(self, completed: List[str]):
        """(winner, top1, top2, conf) over completed-answer pool."""
        if not completed:
            return None, 0, 0, 0.0
        winner, top1, top2, _ = _vote_stats(completed)
        conf = _beta_majority_confidence(top1, top2)
        return winner, top1, top2, conf

    def _update_ema(self, ema_prev: float, new_val: float) -> float:
        """EMA update: ema = (1 - alpha) * ema_prev + alpha * new_val."""
        return (1.0 - self.ema_alpha) * ema_prev + self.ema_alpha * new_val

    def _classify_branch(
        self,
        br: Dict[str, Any],
        pool_winner,
        warm_enough: bool,
    ) -> str:
        """
        Return 'aligned', 'deviant', or 'neutral'.
        - 'aligned'  : has pool_winner and latest_ans == pool_winner
        - 'deviant'  : has pool_winner and latest_ans != pool_winner
        - 'neutral'  : no pool_winner, or warm_up not yet reached
        """
        if not warm_enough or pool_winner is None:
            return "neutral"
        if br["latest_ans"] == pool_winner:
            return "aligned"
        return "deviant"

    def _probe_branch(
        self,
        question,
        br: Dict[str, Any],
        completed_answers: List[str],
        n_steps: int,
    ) -> None:
        """Probe branch br for up to n_steps steps; record completions."""
        for _ in range(n_steps):
            if br["finished"]:
                break
            out = _safe_probe_more(question, br["index"])
            if out is None:
                br["finished"] = True
                if br["latest_ans"] is not None:
                    completed_answers.append(br["latest_ans"])
                break
            new_ans, is_finish = out
            br["probe_count"] += 1
            br["latest_ans"] = new_ans
            br["finished"] = is_finish
            if is_finish:
                completed_answers.append(new_ans)
                break

    # ------------------------------------------------------------------
    # Main solve
    # ------------------------------------------------------------------

    def solve(self, question) -> Optional[str]:
        self._reset_trace()
        self._trace_step(
            event="start",
            goal="initialize CMC run",
            step_input={"beta": self._beta},
            step_output="initialized",
            state={
                "n_init":           self.n_init,
                "max_branch_use":   self.max_branch_use,
                "warm_up":          self.warm_up,
                "abandon_patience": self.abandon_patience,
                "T_ema":            self.T_ema,
                "ema_alpha":        self.ema_alpha,
                "conf_thresh":      self.conf_thresh,
                "delta_slack":      self.delta_slack,
                "burst_aligned":    self.burst_aligned,
                "widen_burst":      self.widen_burst,
                "trend_thresh":     self.trend_thresh,
                "min_complete":     self.min_complete,
            },
            decision="start confidence momentum controller",
        )

        # Branch state:
        #   index          : stable branch_index from probe_new
        #   latest_ans     : current answer (intermediate or final)
        #   finished       : bool — branch exhausted its full budget
        #   abandoned      : bool — dropped due to persistent deviance
        #   probe_count    : number of probe_more steps received
        #   disagree_rounds: consecutive rounds where answer != pool_winner
        branches: List[Dict[str, Any]] = []
        completed_answers: List[str] = []
        total_spawned = 0

        # ---- Phase 0: open n_init branches ----
        for _ in range(self.n_init):
            out = _safe_probe_new(question)
            if out is None:
                break
            ans, idx, is_finish = out
            total_spawned += 1
            br: Dict[str, Any] = {
                "index":           idx,
                "latest_ans":      ans,
                "finished":        is_finish,
                "abandoned":       False,
                "probe_count":     0,
                "disagree_rounds": 0,
            }
            branches.append(br)
            if is_finish:
                completed_answers.append(ans)

        self._trace_step(
            event="init_branches",
            goal="open initial branch batch",
            step_input={"n_init": self.n_init},
            step_output={
                "n_spawned":   total_spawned,
                "n_completed": len(completed_answers),
            },
            state={"total_spawned": total_spawned},
            decision="proceed to main loop",
        )

        if not branches:
            self._trace_step(
                event="finish",
                goal="return final answer",
                step_input={},
                step_output={"answer": None, "stop_reason": "no_branches"},
                state={"total_spawned": 0},
                decision="no branches available",
            )
            return None

        # EMA state — initialised to 0 (no evidence yet)
        ema_conf       = 0.0
        ema_conf_prev  = 0.0
        ema_history: List[float] = []   # last T_ema values for delta computation

        outer_step = 0

        while outer_step < self._MAX_OUTER:

            # ---- Compute current pool stats ----
            pool_winner, top1, top2, pool_conf = self._pool_stats(completed_answers)
            n_complete = len(completed_answers)
            warm_enough = (outer_step >= self.warm_up)

            # ---- Update EMA ----
            ema_conf_prev = ema_conf
            ema_conf = self._update_ema(ema_conf, pool_conf)
            ema_history.append(ema_conf)
            if len(ema_history) > self.T_ema:
                ema_history.pop(0)

            # EMA delta: difference between current EMA and the oldest in window
            if len(ema_history) >= 2:
                ema_delta = ema_history[-1] - ema_history[0]
            else:
                ema_delta = 0.0

            # ---- Classify branches and update disagree_rounds ----
            if warm_enough and pool_winner is not None:
                for br in branches:
                    if br["abandoned"] or br["finished"]:
                        continue
                    tier = self._classify_branch(br, pool_winner, warm_enough)
                    if tier == "deviant":
                        br["disagree_rounds"] += 1
                    else:
                        br["disagree_rounds"] = 0

            # ---- Abandon persistently deviant branches (keep >= 2 alive) ----
            abandoned_this: List[int] = []
            if warm_enough and pool_winner is not None:
                n_alive = sum(
                    1 for br in branches
                    if not br["abandoned"] and not br["finished"]
                )
                cands = sorted(
                    [
                        br for br in branches
                        if not br["abandoned"]
                        and not br["finished"]
                        and br["disagree_rounds"] >= self.abandon_patience
                    ],
                    key=lambda b: -b["disagree_rounds"],
                )
                max_abandon = max(0, n_alive - 2)
                for br in cands[:max_abandon]:
                    br["abandoned"] = True
                    abandoned_this.append(br["index"])

            # ---- Prioritised depth allocation ----
            # Collect active (unfinished, non-abandoned) branches
            active_brs = [
                br for br in branches
                if not br["abandoned"] and not br["finished"]
            ]
            # Sort by probe_count descending (most-invested first)
            active_brs_sorted = sorted(active_brs, key=lambda b: -b["probe_count"])

            probed_this: int = 0
            for br in active_brs_sorted:
                tier = self._classify_branch(br, pool_winner, warm_enough)
                n_steps = self.burst_aligned if tier == "aligned" else 1
                self._probe_branch(question, br, completed_answers, n_steps)
                probed_this += n_steps

            # Recompute pool stats after probing
            pool_winner, top1, top2, pool_conf = self._pool_stats(completed_answers)
            n_complete = len(completed_answers)

            # ---- Refresh EMA after in-round probing completes ----
            ema_conf = self._update_ema(ema_conf, pool_conf)
            if ema_history:
                ema_history[-1] = ema_conf   # update the entry we added this round
            if len(ema_history) >= 2:
                ema_delta = ema_history[-1] - ema_history[0]
            else:
                ema_delta = 0.0

            n_active = sum(
                1 for br in branches if not br["abandoned"] and not br["finished"]
            )

            self._trace_step(
                event="forward",
                goal="probe with priority scheduling + update EMA",
                step_input={
                    "outer_step":  outer_step,
                    "pool_winner": pool_winner,
                    "pool_conf":   round(pool_conf, 4),
                },
                step_output={
                    "n_complete":    n_complete,
                    "n_active":      n_active,
                    "probed_this":   probed_this,
                    "ema_conf":      round(ema_conf, 4),
                    "ema_delta":     round(ema_delta, 4),
                    "abandoned_now": abandoned_this,
                },
                state={"total_spawned": total_spawned},
                decision="evaluate momentum gate and widening",
            )

            # ---- EMA momentum stopping gate ----
            gate_eligible = (
                warm_enough
                and n_complete >= self.min_complete
            )
            # Gate: EMA level is high AND EMA is not actively declining
            gate_fires = (
                gate_eligible
                and ema_conf >= self.conf_thresh
                and ema_delta >= -self.delta_slack
            )

            self._trace_step(
                event="terminate_check",
                goal="EMA momentum gate evaluation",
                step_input={
                    "outer_step":   outer_step,
                    "conf_thresh":  self.conf_thresh,
                    "delta_slack":  self.delta_slack,
                    "min_complete": self.min_complete,
                    "warm_up":      self.warm_up,
                },
                step_output={
                    "ema_conf":      round(ema_conf, 4),
                    "ema_delta":     round(ema_delta, 4),
                    "pool_conf":     round(pool_conf, 4),
                    "n_complete":    n_complete,
                    "gate_eligible": gate_eligible,
                    "gate_fires":    gate_fires,
                },
                state={"total_spawned": total_spawned},
                decision="stop if EMA gate fires",
            )

            if gate_fires:
                self._trace_step(
                    event="finish",
                    goal="return final answer",
                    step_input={"outer_step": outer_step},
                    step_output={
                        "answer":      pool_winner,
                        "stop_reason": "ema_momentum_gate",
                        "ema_conf":    round(ema_conf, 4),
                        "ema_delta":   round(ema_delta, 4),
                        "n_complete":  n_complete,
                    },
                    state={"total_spawned": total_spawned},
                    decision="EMA level high + momentum non-negative",
                )
                return pool_winner

            # ---- All branches resolved? ----
            all_resolved = all(br["finished"] or br["abandoned"] for br in branches)
            if all_resolved:
                break

            # ---- Confidence-trend widening ----
            # Widen when EMA trend is weak (plateau/regression) AND budget remains.
            # Confidence growing strongly -> no need to widen (depth alone will converge).
            can_widen = (
                total_spawned < self.max_branch_use
                and total_spawned < self._MAX_BRANCH
            )
            trend_weak = ema_delta <= self.trend_thresh
            want_widen = (
                can_widen
                and trend_weak
                and outer_step >= max(1, self.warm_up // 2)
                and ema_conf < self.conf_thresh
            )

            spawned_now = 0
            if want_widen:
                for _ in range(self.widen_burst):
                    if total_spawned >= self.max_branch_use:
                        break
                    if total_spawned >= self._MAX_BRANCH:
                        break
                    out = _safe_probe_new(question)
                    if out is None:
                        break
                    ans, idx, is_finish = out
                    total_spawned += 1
                    spawned_now += 1
                    br_new: Dict[str, Any] = {
                        "index":           idx,
                        "latest_ans":      ans,
                        "finished":        is_finish,
                        "abandoned":       False,
                        "probe_count":     0,
                        "disagree_rounds": 0,
                    }
                    branches.append(br_new)
                    if is_finish:
                        completed_answers.append(ans)

            self._trace_step(
                event="update_states",
                goal="confidence-trend widening snapshot",
                step_input={
                    "outer_step":   outer_step,
                    "want_widen":   want_widen,
                    "ema_conf":     round(ema_conf, 4),
                    "ema_delta":    round(ema_delta, 4),
                    "trend_thresh": self.trend_thresh,
                },
                step_output={
                    "spawned_now":   spawned_now,
                    "total_spawned": total_spawned,
                    "all_resolved":  all_resolved,
                },
                state={"n_active": n_active},
                decision="continue main loop",
            )

            outer_step += 1

        # ---- Final answer ----
        final_winner, _, _, final_conf = self._pool_stats(completed_answers)
        if final_winner is None:
            all_latest = [
                br["latest_ans"]
                for br in branches
                if not br["abandoned"] and br["latest_ans"] is not None
            ]
            final_winner = _majority_answer(all_latest)
            final_conf = 0.0

        self._trace_step(
            event="finish",
            goal="return final answer",
            step_input={"outer_step": outer_step},
            step_output={
                "answer":        final_winner,
                "stop_reason":   "loop_end",
                "ema_conf":      round(ema_conf, 4),
                "pool_conf":     round(final_conf, 4),
                "n_complete":    len(completed_answers),
                "total_spawned": total_spawned,
            },
            state={"total_spawned": total_spawned},
            decision="majority of completed answers at loop end",
        )
        return final_winner
\end{Verbatim}
\end{tcolorbox}